\documentclass[10pt,twocolumn,letterpaper]{article}

\usepackage[pagenumbers]{wacv} 

\usepackage{graphicx}
\usepackage{amsmath}
\usepackage{amssymb}

\usepackage{tabularx}
\usepackage{makecell}
\usepackage{booktabs}
\usepackage{bm}
\usepackage{algorithm}
\usepackage{algorithmicx}
\usepackage{algpseudocode}
\usepackage[dvipsnames]{xcolor}
\usepackage{multirow}
\usepackage{pifont}

\newcommand{\cmark}{\ding{51}}%
\newcommand{\xmark}{\ding{55}}%

\usepackage{colortbl}

\usepackage[accsupp]{axessibility}

\usepackage[pagebackref,breaklinks,colorlinks]{hyperref}

\usepackage[capitalize]{cleveref}
\crefname{section}{Sec.}{Secs.}
\Crefname{section}{Section}{Sections}
\Crefname{table}{Table}{Tables}
\crefname{table}{Tab.}{Tabs.}

\def\wacvPaperID{1362} 
\def\confName{WACV}
\def\confYear{2025}

\begin{document}

\title{Enhancing Novel Object Detection via Cooperative Foundational Models}

\author{Rohit Bharadwaj\(^{1}\) \quad Muzammal Naseer\(^{4}\) \quad Salman Khan\(^{1,3}\) \quad Fahad Shahbaz Khan\(^{1,2}\)\\
\(^{1}\)MBZUAI \hspace{1em} \(^{2}\)Linköping University \hspace{1em} \(^{3}\)Australian National University \hspace{1em} \(^{4}\)Khalifa University\\
{\fontsize{8}{10}\selectfont \tt\small rohit.bharadwaj@mbzuai.ac.ae, muhammadmuzammal.naseer@ku.ac.ae,}\\
{\fontsize{8}{10}\selectfont \tt\small salman.khan@mbzuai.ac.ae, fahad.khan@mbzuai.ac.ae}
}

\maketitle

\begin{abstract}
In this work, we address the challenging and emergent problem of novel object detection (NOD), focusing on the accurate detection of both known and novel object categories during inference. Traditional object detection algorithms are inherently closed-set, limiting their capability to handle NOD. We present a novel approach to transform existing closed-set detectors into open-set detectors. This transformation is achieved by leveraging the complementary strengths of pre-trained foundational models, specifically CLIP and SAM, through our cooperative mechanism. Furthermore, by integrating this mechanism with state-of-the-art open-set detectors such as GDINO, we establish new benchmarks in object detection performance. Our method achieves 17.42 mAP in novel object detection and 42.08 mAP for known objects on the challenging LVIS dataset. Adapting our approach to the COCO OVD split, we obtain an impressive result of 49.6 Novel AP50, which outperforms existing SOTA methods with similar backbone. Our code is available at: \\ \href{https://rohit901.github.io/coop-foundation-models/}{https://rohit901.github.io/coop-foundation-models/}
\end{abstract}


\section{Introduction}
\label{sec:intro}

Object detection serves as a cornerstone in computer vision, with broad applications from autonomous driving and robotic vision to video surveillance and pedestrian detection~\cite{od-survey, driving-autono, wang2023multi}. Current state-of-the-art approaches such as Mask-RCNN~\cite{maskrcnn}, and DETR~\cite{detr} operate under a closed-set paradigm, where detection is limited to predefined classes seen during training. This limitation does not align with the dynamic and evolving nature of real-world environments where object classes follow a long-tail distribution, with numerous rare and a few common classes. Collecting resource-intensive datasets to represent these rare classes is an impractical task~\cite{gupta2019lvis}. Thus, the development of open-set detectors that can generalize beyond their training data is crucial for their practical deployment.

The challenge of detecting novel objects is exacerbated by the absence of labeled annotations for such classes. Prior efforts have largely addressed \textit{Generalized Novel Class Discovery (GNCD)}, utilizing semi-supervised and contrastive learning techniques that assume access to pre-cropped objects and balanced datasets. However, these methods are predominantly focused on classification rather than localization, leaving a gap in novel object detection capabilities.

\noindent RNCDL\cite{fomenko2022learning} is a pioneering approach addressing novel object detection (NOD; requires both localization and recognition of known and novel classes) through a training pipeline encompassing both supervised and self-supervised learning. Nonetheless, RNCDL lacks the ability to assign semantic labels to novel categories without additional validation sets, and its performance falls short for practical applications.

\begin{figure}[t]
    \centering
    \begin{subfigure}[b]{0.48\columnwidth}
        \includegraphics[width=\textwidth]{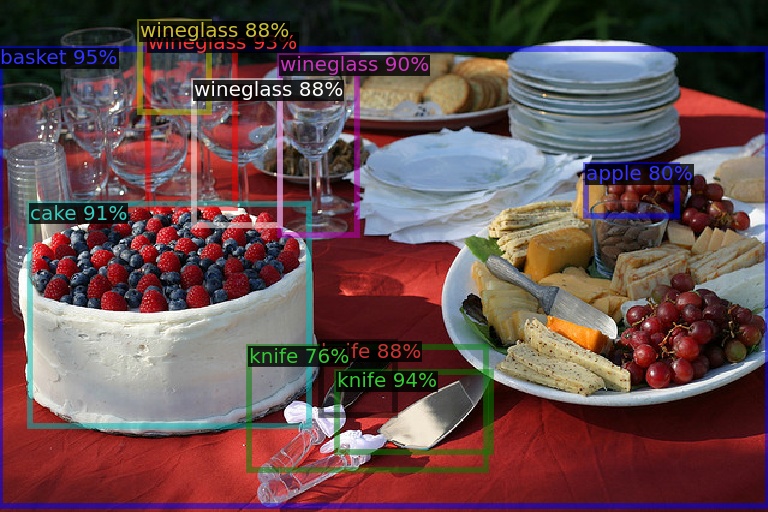}
        \caption*{(a) RNCDL output}
        \label{fig:rncdl_output}
    \end{subfigure}
    \hfill
    \begin{subfigure}[b]{0.48\columnwidth}
        \includegraphics[width=\textwidth]{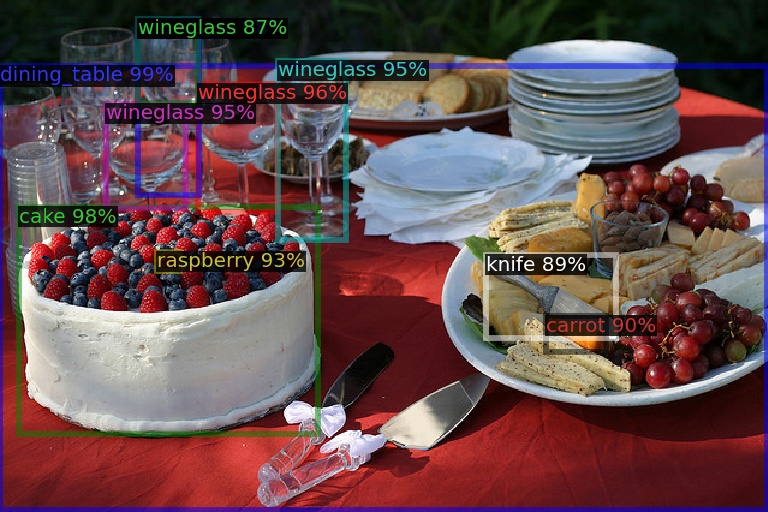}
        \caption*{(b) Our Mask-RCNN method}
        \label{fig:maskrcnn_output}
    \end{subfigure}
    \caption{Comparison of top-10 predictions: (a) RNCDL~\cite{fomenko2022learning} can result in imprecise localization and misclassification (e.g., basket, apple), versus (b) our open-set Mask-RCNN, demonstrating accurate detection and categorization of unique objects in the scene.}
    \label{fig:intro_comparison}
\end{figure}

Emerging approaches in \textit{open-vocabulary object detection (OVD; similar to NOD, but main focus is on novel classes, and requires large-scale training)} have demonstrated promise through the integration of Vision-Language Models (VLMs) like CLIP~\cite{clip}, which leverage language embeddings of category names to generalize detection capabilities. Techniques such as GLIP~\cite{li2021grounded} and Grounding DINO (GDINO)~\cite{groundingdino} further intertwine language and vision modalities at various architectural stages, achieving detection through text inputs. These advancements, however, are contingent upon extensive training and computational resources, implicating significant environmental and financial costs~\cite{patterson2021carbon, strubell2020energy}.

In this work, we present a cooperative mechanism that harnesses the complementary strengths of foundational models such as CLIP and SAM to transition existing closed-set detectors into open-set detectors for novel object detection (NOD). These foundational models, trained on diverse datasets, are adept at generalizing across tasks and distributions unseen during training. Our approach uses CLIP's zero-shot classification in tandem with background boxes identified by Mask-RCNN to determine novel class labels and their confidence scores. These boxes then guide SAM in refining the predictions and eliminating spurious background detections. Ours is the first approach to demonstrate the potential of existing foundational VLMs for NOD.

We further demonstrate that our cooperative mechanism, when paired with open-set detectors like GDINO, elevates performance metrics across both known and novel object categories. Through extensive evaluations on the challenging LVIS~\cite{gupta2019lvis} dataset and the COCO~\cite{coco} open-vocabulary benchmark, our method sets new state-of-the-art, particularly in Novel AP50 metric. Ablation studies highlight the significant contributions of individual and novel components, such as the Synonym Averaged Embedding Generator (SAEG) and the Score Refinement Module (SRM). Our contributions are as follows:
\begin{enumerate}
    \item A cooperative mechanism leveraging the strengths of pre-trained foundational models like CLIP and SAM, transforming closed-set detectors into proficient open-set detectors for novel object detection.
    \item A modular design enabling integration with state-of-the-art open-set detectors to enhance Known and Novel class AP in challenging settings.
    \item We demonstrate the effectiveness of our approach compared to existing state-of-the-art in open-set, unknown, and open-vocabulary object detection by various experiments conducted on challenging datasets like COCO and LVIS. Further, our ablation studies demonstrate the strength of proposed approach and its constituent elements.
\end{enumerate}

\begin{figure}[!t]
    \centering
    \includegraphics[width=0.9\columnwidth]{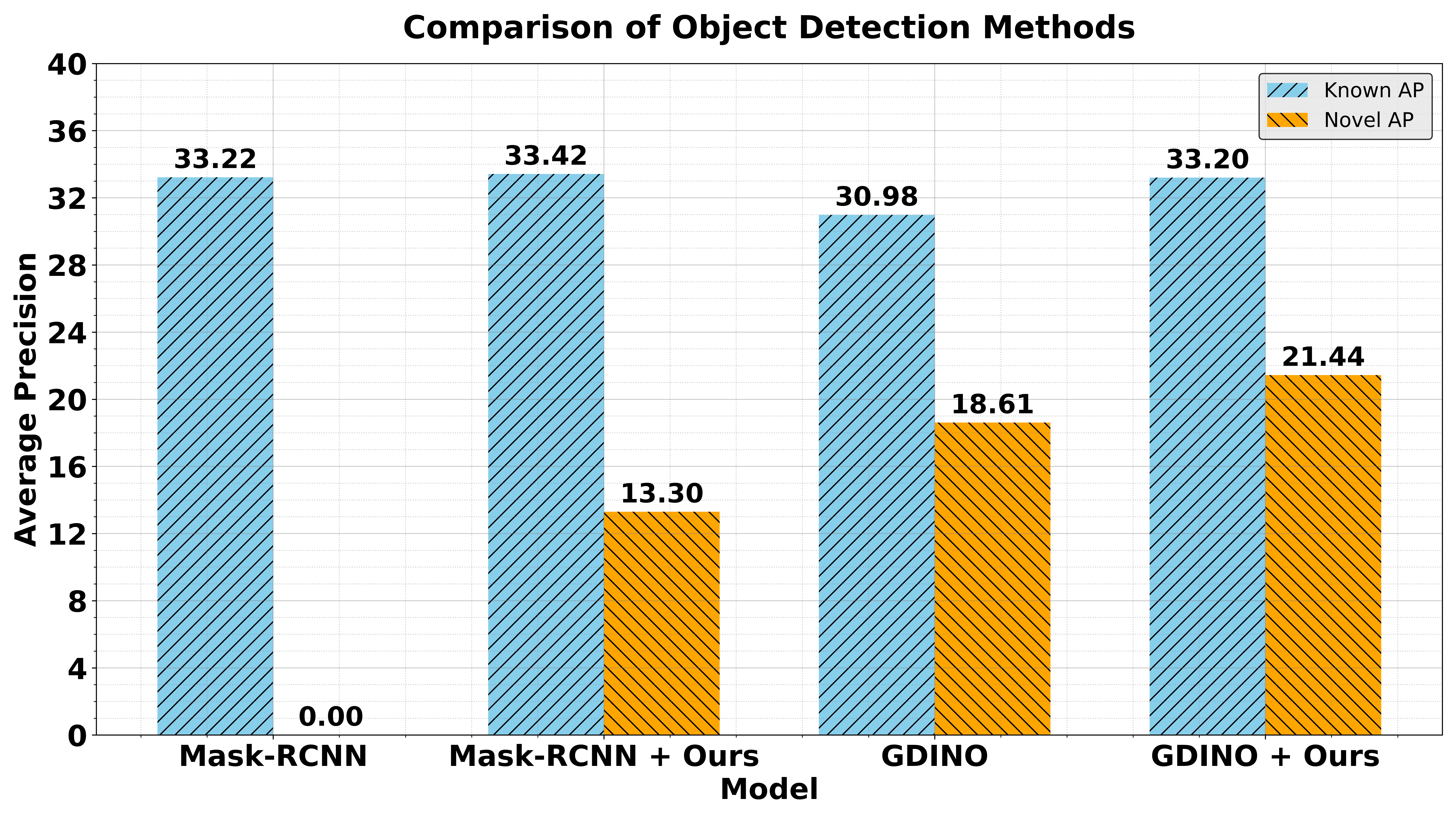}
    \caption{\small Comparative analysis of object detection methods on \texttt{lvis\_v1\_val\_subset} dataset. The closed-set Mask-RCNN does not detect novel classes, however, the performance consistently improves when combined with our cooperative mechanism integrating different foundational models.
    }
    \label{fig:rcnn-gdino-openset-comp}
\end{figure}

\section{Related Work}

Recent research in object detection has made significant strides, yet the challenge of detecting objects outside the predefined classes remains unsolved. Existing methods often assume a semi-supervised setting and focus on classification rather than the complex task of simultaneous classification and localization, which our work tackles. GCD~\cite{vaze2022generalized} exploits the capabilities of Vision Transformers and contrastive learning to identify novel classes, avoiding the overfitting issues of parametric classifiers. OpenLDN~\cite{rizve2022openldn} adopts a similar goal, leveraging image similarities and bi-level optimization to cluster novel class instances, while PromptCAL~\cite{zhang2023promptcal} introduces contrastive affinity learning with visual prompts to enhance class discovery.
In the realm of unknown object detection, methods such as VOS~\cite{du2022vos} synthesize virtual outliers to differentiate between known and unknown objects. UnSniffer~\cite{liang2023unknown}, the current leading method, uses a generalized object confidence score and an energy suppression loss to distinguish non-object background samples. For Open-Vocabulary Detection (OVD), approaches like ViLD~\cite{gu2022openvocabulary}, OV-DETR\cite{ov-detr}, and BARON~\cite{wu2023aligning} employ knowledge distillation from Vision-Language Models to align region embeddings with VLM features. The state-of-the-art CORA\cite{wu2023cora} further refines this by combining region prompting with anchor pre-matching in a DETR-based framework. RNCDL~\cite{fomenko2022learning} also addresses novel object detection, but its two-stage training pipeline, reliance on a validation set for semantic label assignment, and poor performance limit its practicality. Unlike RNCDL, our work is able to achieve superior performance in a training-free manner.
Lastly, methods like GLIP~\cite{li2021grounded} and GDINO~\cite{groundingdino} harness natural language to expand detection capabilities. However, the extensive training required for these methods incurs significant financial and environmental costs. Our approach, on the other hand, leverages complementary strengths of pre-trained foundational models to achieve superior performance without the need for extensive training, showcasing a more practical and accessible solution for novel object detection. Further, the modular nature of our cooperative mechanism allows it to be combined with any existing state-of-the-art open-set detectors like GDINO to further improve the overall performance across both known and novel classes.

\section{Novel Object Detection}
\begin{figure*}[!t]
    \centering
    \includegraphics[width=0.75\textwidth]{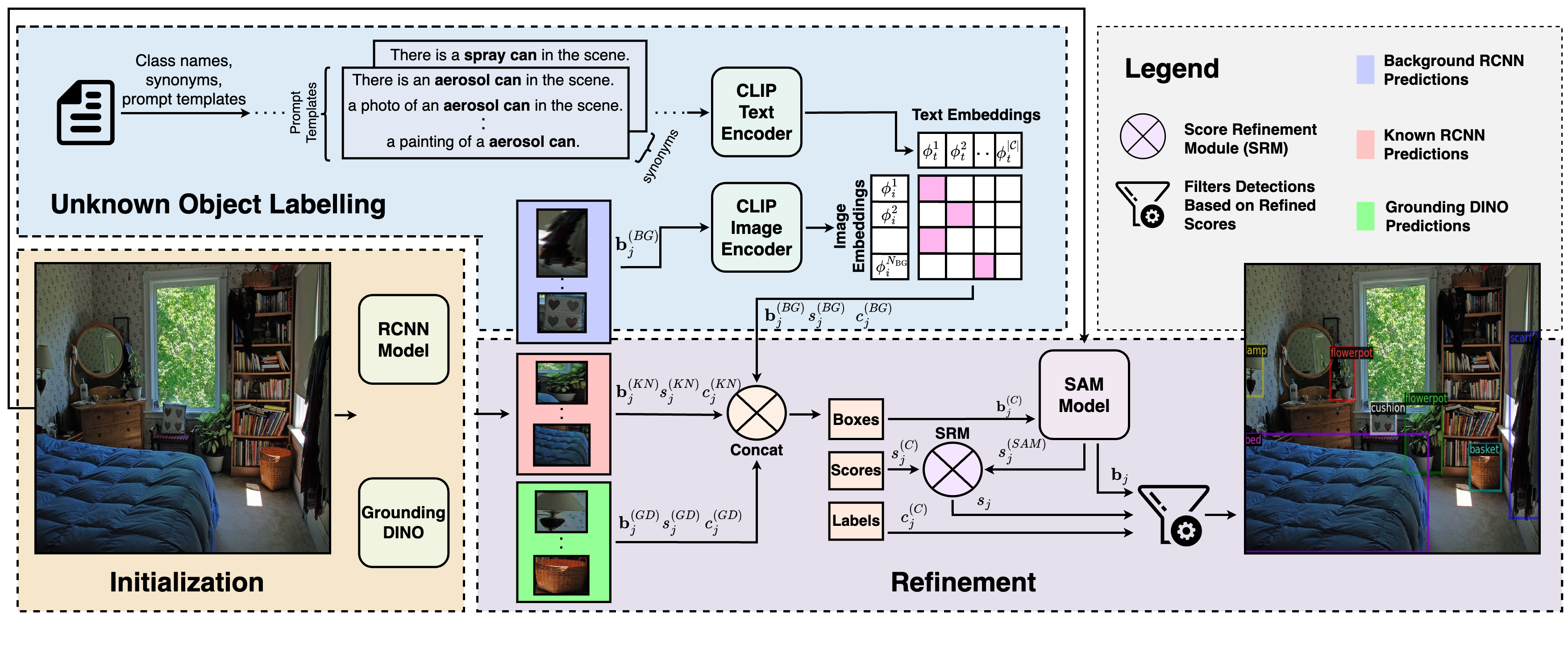}
    \caption{Our proposed cooperative mechanism integrates pre-trained foundational models such as CLIP, SAM, and GDINO with a Mask-RCNN model in order to identify and semantically label both known and novel objects. These foundational model interacts using different components including Initialization, Unknown Object Labelling, and Refinement to refine and categorize objects.}
    \label{fig:architecture}
\end{figure*}

Novel Object Detection (NOD) deals with object detection under heterogeneous object distributions during training and inference. Specifically, we assume that the distribution of classes observed during training might differ from the distribution observed during inference. As a result, we may encounter both known and previously unknown object classes. Our objective is to identify objects from known and novel categories during inference while assigning semantic labels to each object. 

Formally, we define $\mathcal{D}_{\text{train}} = \{(\mathbf{x}_i, y_i)\}_{i=1}^{N} \in \mathcal{X} \times \mathcal{Y}_{\text{train}}$ as the training dataset. Each $\mathbf{x}_i \in \mathbb{R}^{3 \times h \times w}$ represents an input image and $y_i = \{(\mathbf{b}_j, c_j)\}_{j=1}^{L}$ comprises bounding box label $\mathbf{b}_j \in \mathbb{R}^4$ and class label $c_j \in \mathcal{C}^{\text{known}}$. We represent the test dataset as $\mathcal{D}_{\text{test}} = \{(\mathbf{x}_i, y_i)\}_{i=1}^{M} \in \mathcal{X} \times \mathcal{Y}_{\text{test}}$, where $\mathcal{Y}_{\text{test}} \supset \mathcal{Y}_{\text{train}}$. Importantly, class labels during testing, $c_j$, can belong to an extended set $\mathcal{C} \supset \mathcal{C}^{\text{known}}$.

Novel classes are by definition unknown; therefore, it is challenging to detect them during inference. The conventional object detection models such as Mask-RCNN \cite{maskrcnn} and DETR \cite{detr} are examples of closed-set detectors unable to generalize beyond the classes of their training data. These models are unsuitable for detecting novel objects. We therefore require open-set object detectors to solve the problem of NOD. Recently, Grounding DINO \cite{groundingdino} (GDINO) achieved open-set generalization by introducing natural language to a closed-set detector DINO \cite{dino}. The model processes natural language input in the form of category names or classes to detect arbitrary objects within an image.
The key to the open-set success of GDINO is language and vision fusion along with large-scale training.

\textbf{Approach Overview.} In this work, we show how to convert an existing closed-set detector, i.e, pre-trained Mask-RCNN, to an open-set detector by utilizing the complementary strengths of pre-trained foundational models such as CLIP~\cite{clip}, and SAM~\cite{sam} via our \emph{cooperative mechanism}. Our proposed mechanism leverages CLIP's understanding of unseen classes with Mask-RCNN's ability to localize background objects for finding novel object classes (\cref{subsec:Cooperative Foundational Models for NcD}). The bounding boxes are then refined by using SAM's instance mask-to-box capabilities (\cref{alg:scoreRefinement}). We observe that the novel class AP of Mask-RCNN is zero due to its closed-set nature (\cref{fig:rcnn-gdino-openset-comp}). Our cooperative mechanism, however, can be applied to the same closed-set Mask-RCNN model and result in notable gains in novel AP, as well as better performance in known AP compared to GDINO and the baseline Mask-RCNN. Additionally, when our proposed cooperative mechanism is combined with open-set detectors like GDINO, we observe overall performance gains (\cref{fig:rcnn-gdino-openset-comp}).

Our approach uses off-the-shelf pre-trained models in three stages; a) Initialization, b) Unknown Object Labelling, and c) Refinement. The initialization stage consists of getting bounding boxes of known and unknown object categories from complementary off-the-shelf detectors (Mask-RCNN \cite{maskrcnn}, Grounding DINO~\cite{groundingdino}). The second stage involves processing unlabelled background bounding boxes by a pre-trained language-image model CLIP \cite{clip}. Finally, the detected boxes are refined using SAM \cite{sam}. We first provide an overview of CLIP, SAM, and GDINO in the next \cref{subsec:Preliminary}. 

\subsection{Preliminaries}
\label{subsec:Preliminary}
\textbf{Contrastive Language-Image Pre-training (CLIP).} The CLIP model incorporates a dual-encoder architecture tailored for multi-modal learning, where the text and image encoders are represented by \( \mathcal{F}_{t}^{(\text{CLIP})} \) and \(\mathcal{F}_{i}^{(\text{CLIP})}\), respectively. Training involves a batch of $N$ image-text pairs, where the goal is to align the image embeddings \( \bm{\phi}_{i} \in \mathbb{R}^{d}\) with their corresponding text embeddings \(\bm{\phi}_{t} \in \mathbb{R}^{d}\). 

For zero-shot classification, given an image \(\mathbf{x} \) and a set of \( \lvert \mathcal{C} \rvert \) unique classes, we generate \( \lvert \mathcal{C} \rvert \) textual prompts using a template \(T(\cdot)\) — for instance, “a photo of a [CLASS].” By replacing the [CLASS] token with each class name \(c\) from our targeted dataset, we form the textual descriptors. The embeddings for these descriptors, \( \bm{\Phi}_{t} = \mathcal{F}_{t}^{(\text{CLIP})}\left(T(\mathcal{C})\right) \), and for the image \( \bm{\phi}_{i} = \mathcal{F}_{i}^{(\text{CLIP})}\left(\mathbf{x}\right) \), are used to compute logits as:
$\bm{\phi}_{i}^{T} \bm{\Phi}_{t} \in \mathbb{R}^{1 \times \lvert \mathcal{C} \rvert}$.
The predicted class for the image is the one with the highest corresponding score.

\noindent\textbf{Segment Anything Model (SAM)}. SAM is a recent foundational model for generic class-agnostic image segmentation. The model is comprised of three primary modules: an image encoder \( \mathcal{F}_{i}^{(\text{SAM})} \), a prompt encoder \( \mathcal{F}_{p}^{(\text{SAM})} \), and a mask decoder \( \mathcal{G}_{m}^{(\text{SAM})} \). Formally, for an input image \( \mathbf{x} \), and a set of prompts \( \mathcal{P} = \{\mathbf{p}_1, ..., \mathbf{p}_M\} \), SAM generates refined segmentation masks \( \{\mathbf{m}_1, ..., \mathbf{m}_M\} \) and a corresponding set of confidence scores \( \mathbf{s}^{(\text{SAM})} = \{s_1^{(\text{SAM})}, ..., s_M^{(\text{SAM})}\} \). The prompts \( \mathcal{P} \) can include any combination of points, bounding boxes, or rough masks. These are encoded through \( \mathcal{F}_{p}^{(\text{SAM})} \) to produce prompt embeddings \( \bm{\phi}_{p} \). Concurrently, \( \mathcal{F}_{i}^{(\text{SAM})} \) processes \( \mathbf{x} \), yielding image embeddings \( \bm{\phi}_{i} \). These two embeddings are then fused and passed to \( \mathcal{G}_{m}^{(\text{SAM})} \), for refined segmentation masks \( \mathbf{m} \) and their scores \( \mathbf{s}^{(\text{SAM})} \).

\noindent \textbf{Grounding DINO (GDINO)}. GDINO utilizes the DETR~\cite{detr} architecture for a transformer-based framework for object detection. It adopts the Swin Transformer~\cite{swintrans} as the visual backbone and BERT~\cite{bert} for the textual encoding. In this setup, given a pair consisting of an image \( \mathbf{x} \) and a concatenated textual input of class names from the set of all classes \( \mathcal{C} \), the model yields a fixed number (\( \texttt{num\_query}\)\,=\,900) of predicted bounding boxes \( \{\mathbf{b}_{i}^{(\text{GD})}\}_{i=1}^{\texttt{num\_query}} \). Alongside, it generates a set of confidence scores \( \{s_{i}^{(\text{GD})}\}_{i=1}^{\texttt{num\_query}} \) and associates each detected region with a class label \( \{c_{i}^{(\text{GD})}\in \mathcal{C}\}_{i=1}^{\texttt{num\_query}}  \).

\subsection{Cooperative Foundational Models for NOD}
\label{subsec:Cooperative Foundational Models for NcD}

\textbf{Initialization}
\label{subsec:Initialization}
In our proposed pipeline, the first stage involves initializing bounding boxes using off-the-shelf detectors, such as Mask-RCNN and GDINO, to obtain the unrefined bounding boxes from the input image. The outputs of the detectors based on a conventional two-stage RPN design are combined with a complementary DETR architecture to compensate for the weaknesses of each component. For example, Mask-RCNN does not integrate linguistic cues, and DETR has only a limited set of predicted bounding boxes (i.e. \texttt{num\_query}).

Specifically, the Mask-RCNN is a unimodal, non-transformer based two-stage model, hence it does not provide the flexibility to detect objects from an open-vocabulary. The RCNN family of object detection models are known as two-stage detectors because, in the first stage, they output all the object proposals in the given image, which may include known as well as unidentifiable background objects, and in the second stage, the model generates the final box predictions based on the initial set of proposals. In the absence of class labels for novel objects within a dataset, these categories are generally classified as \textit{background} class. To solve the NOD problem, the Mask-RCNN model needs to output bounding boxes for background classes as well as known classes.
Similarly, due to the \texttt{num\_query} bounding box limitation, the DETR based GDINO model does not perform well on rare classes e.g., within the LVIS dataset~\cite{groundingdino}.
As shown in \cref{fig:architecture}, GDINO misses out on detecting some of the other objects like \textit{cushion}, and \textit{scarf} which are present in the input image.

Formally, for a given input image, \( \mathbf{x} \), we obtain the outputs from GDINO - specifically, the bounding boxes \( \mathbf{b}_{j}^{(\text{GD})} \), class confidence scores \( s_{j}^{(\text{GD})} \), and the associated class label \( c_{j}^{(\text{GD})} \in \mathcal{C} \), where \( j = 1, \ldots, N_{\text{GD}}\). For the same input image \( \mathbf{x} \), we also get the Mask-RCNN outputs which include the known bounding boxes; \( \mathbf{b}_{j}^{(\text{KN})} \), class confidence scores \( s_{j}^{(\text{KN})} \), and class labels \( c_{j}^{(\text{KN})} \in \mathcal{C}^{\text{known}} \), where \( j = 1, \ldots, N_{\text{KN}} \), and the background bounding boxes; bounding boxes \( \mathbf{b}_{j}^{(\text{BG})} \), where \( j = 1, \ldots, N_{\text{BG}} \). Since the closed-set Mask-RCNN is not able to output the class labels and confidence scores for the background boxes, we describe our approach utilizing CLIP to obtain this missing data, and to convert the closed-set Mask-RCNN model to an open-set detector in the next subsection.

\textbf{Unknown Object Labelling}
\label{subsec:Unknown Object Labelling}
To convert the existing closed-set Mask-RCNN to an open-set detector, we utilize the zero-shot capabilities of a vision language model such as CLIP to obtain the class labels and confidence scores for the background boxes \(\mathbf{b}_{j}^{(\text{BG})}\). First, the regions of interest (ROIs) are cropped from the raw images, $\mathbf{x}_i$, using the background boxes, $\mathbf{b}_{j}^{(BG)}$. We obtain \(N_{\text{BG}}\) number of ROIs that serve as the image inputs for zero-shot classification with CLIP. As shown in \cref{fig:architecture}, we obtain the visual embedding for these ROIs denoted as $\{\bm{\phi}_{i}^{k}\}_{k=1}^{N_{BG}}$. 

\textbf{Synonym Averaged Embedding Generator (SAEG).}
To obtain the text embeddings corresponding to each class $c \in \mathcal{C}$, we define \( \mathcal{T} \) as set of all prompt templates, \( \mathcal{S}_i \) as set of all synonyms for the \(i\)-th class \( C_i \) in the dataset and \( \mathcal{C} \) as set of all classes, \( \{C_1, C_2, \ldots, C_{\lvert \mathcal{C} \rvert} \} \). Then for each synonym \( s \) in \( \mathcal{S}_i \), we proceed as follows. First, the set of prompted texts for the synonym \( s \) is generated as \( \mathcal{P}_s = \{ T(s) : T \in \mathcal{T} \} \).
Afterwards, these prompted texts are tokenized and encoded using CLIP to obtain their embeddings, i.e, \( \bm{\phi}_{t}^{\text{syn}} = \mathcal{F}_{t}^{(\text{CLIP})}\left( \mathcal{P}_{s} \right) \in \mathbb{R}^{d \times n} \), where \( n = \lvert \mathcal{P}_s \rvert \).
The resulting embeddings are normalized and averaged to obtain the representative text feature \( \mathbf{f}_{s} \in \mathbb{R}^d \) for the synonym \( s \).

    \begin{equation}
        \mathbf{f}_{s} = \frac{1}{\lvert \mathcal{P}_s \rvert} \sum_{\mathbf{e} \in \bm{\phi}_{t}^{\text{syn}}} \frac{\mathbf{e}}{\lVert \mathbf{e} \rVert}.
    \end{equation}
The text features for all synonyms of a class are averaged to generate the text feature \( \bm{\phi}_{t}^{i} \in \mathbb{R}^d \) for that class \( C_i \).
\begin{equation}
        \bm{\phi}_{t}^{i} = \frac{1}{\lvert \mathcal{S}_i \rvert} \sum_{\mathbf{f}_s \in \mathcal{S}_i} \frac{\mathbf{f}_s}{\lVert \mathbf{f}_s \rVert}.
    \end{equation}

The final text feature matrix \( \bm{\Phi}_{t} \) for all possible classes can then be represented as:
\begin{equation}
    \bm{\Phi}_{t} = [\bm{\phi}_{t}^{1}, \bm{\phi}_{t}^{2}, \ldots, \bm{\phi}_{t}^{\lvert \mathcal{C} \rvert}] \in \mathbb{R}^{d \times \lvert \mathcal{C} \rvert}.
\end{equation}

Cosine similarity between these image and enriched text embeddings is computed to finalize class predictions and their confidence scores for the background boxes.

\textbf{Refinement}
\label{subsec:Refinement}
We consolidate bounding boxes, confidence scores, and class labels obtained from Mask-RCNN and Grounding DINO models. Specifically, we concatenate the outputs as depicted in \cref{fig:architecture} to formulate a unified set of bounding boxes, $\mathbf{b}_{j}^{(\text{C})}$, confidence scores, $s_{j}^{(\text{C})}$, and class labels, $c_{j}^{(\text{C})}$, where $j = 1, \ldots, N_{\text{C}}$. Finally, the total number of combined bounding boxes are $ N_{\text{C}} = N_{\text{KN}} + N_{\text{BG}} + N_{\text{GD}}$.

This combined set of bounding boxes, $\mathbf{b}_{j}^{(\text{C})}$, is inherently noisy. To solve this problem and to achieve robust generalization beyond the training data distribution, we utilize SAM~\cite{sam}, which allows for efficient zero-shot generalization. The raw image $\mathbf{x}_{i}$, and the set of prompts \(\mathcal{P} = \mathbf{b}_{j}^{(C)} \), serve as inputs to SAM. The output is a set of refined segmentation masks for each of the prompted bounding boxes. From these improved masks, we extract new bounding boxes, constituting our final set of refined bounding box predictions, $\mathbf{b}_{j}$. Subsequently, we introduce the Score Refinement Module (SRM). This novel component integrates confidence scores obtained from SAM with the combined confidence scores, $s_{j}^{(\text{C})}$, to filter and re-weight the high-quality object predictions, which significantly improves the overall object detection performance as explained next.

\textbf{Score Refinement Module (SRM).}
Our Score Refinement Module (SRM) integrates two sets of scores: mask-quality scores, \(s_{j}^{(\text{SAM})}\), and combined confidence scores \(s_{j}^{(\text{C})}\). Both sets of scores are individually subjected to MinMax standardization~\cite{scikit-learn}. The final refined confidence scores, \(s_{j}\), are obtained by element-wise multiplication of the standardized scores. Algorithm~\ref{alg:scoreRefinement} describes our SRM for score refinement. Finally, using the refined scores, $s_{j}$, refined bounding boxes, $\mathbf{b}_{j}$, and the combined class labels, $c_{j}^{(\text{C})}$, we filter high-quality object predictions.

\begin{algorithm}[!t]
\caption{Score Refinement Process}
\label{alg:scoreRefinement}
\begin{algorithmic}[1]
\State \textbf{Input:} \parbox[t]{.85\linewidth}{Combined scores \(s_{j}^{(\text{C})}\), SAM scores \(s_{j}^{(\text{SAM})}\)}
\State \textbf{Output:} Refined scores \(s_{j}\)
\State Initialize MinMaxScaler (from sk-learn library) for combined scores and SAM scores as \( \text{scaler}, \text{scaler\_sam}\)
\State \parbox[t]{.8\linewidth}{\(s_{j}^{(\text{C})} \leftarrow \text{scaler.fit\_transform}(s_{j}^{(\text{C})})\)}
\State \parbox[t]{.8\linewidth}{\(s_{j}^{(\text{SAM})} \leftarrow \text{scaler\_sam.fit\_transform}(s_{j}^{(\text{SAM})})\)}
\State \(s_{j} = s_{j}^{(\text{C})} \times s_{j}^{(\text{SAM})}\)
\end{algorithmic}
\end{algorithm}

\section{Experiments and Results}
\begin{figure*}[htbp]
    \centering
    \begin{subfigure}[b]{0.22\textwidth}
    \centering \small RNCDL
        \includegraphics[width=\textwidth]{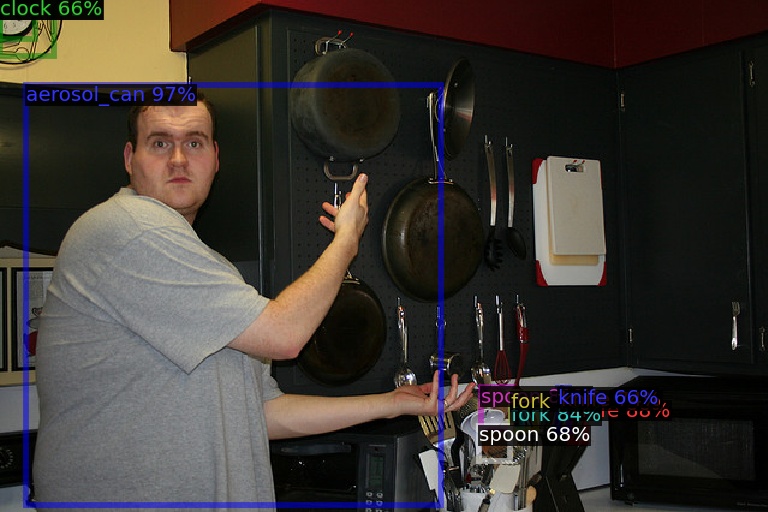}
        \label{fig:1_rncdl_1}
    \end{subfigure}
    \begin{subfigure}[b]{0.22\textwidth}
    \centering \small GDINO
        \includegraphics[width=\textwidth]{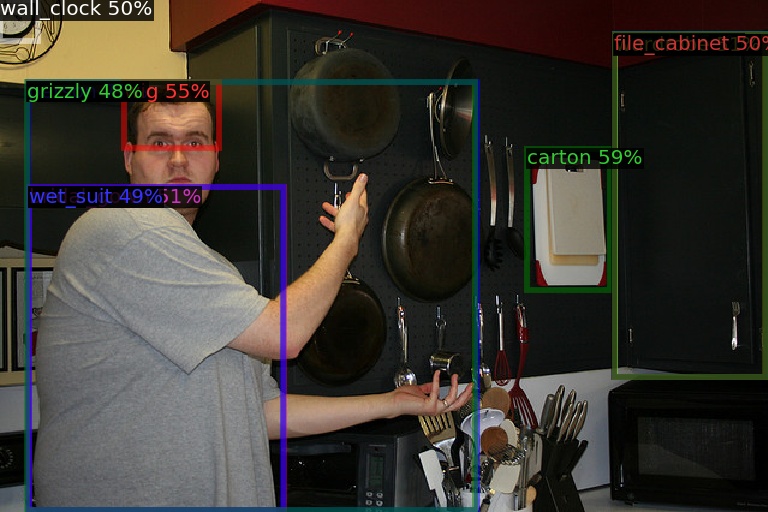}
        \label{fig:1_gdino_1}
    \end{subfigure}
    \begin{subfigure}[b]{0.22\textwidth}
    \centering \small Mask-RCNN \& CLIP
        \includegraphics[width=\textwidth]{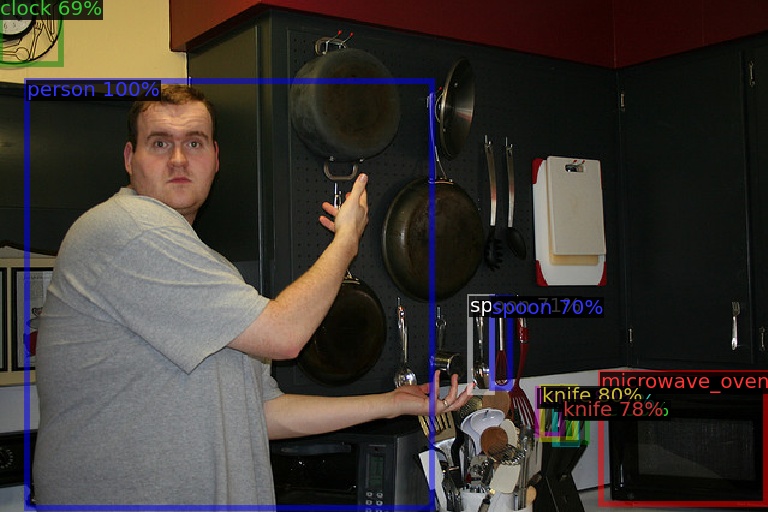}
        \label{fig:1_our_method_1}
    \end{subfigure}
    \begin{subfigure}[b]{0.22\textwidth}
    \centering \small Ours with GDINO \& Mask-RCNN
        \includegraphics[width=\textwidth]{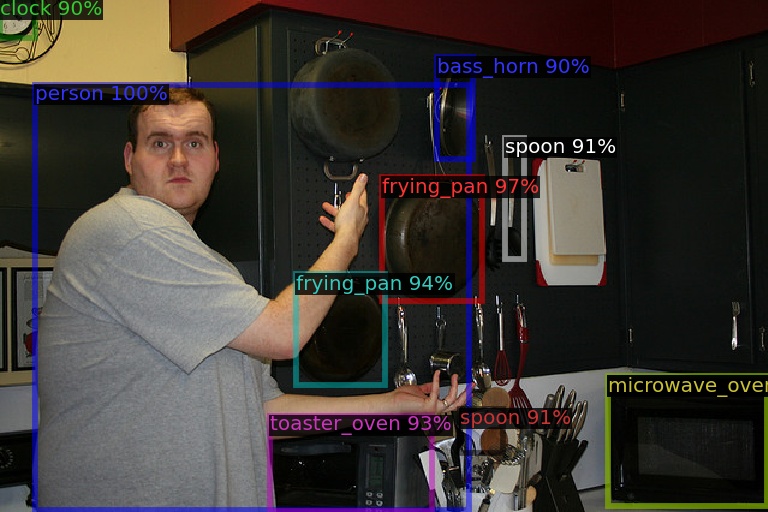}
        \label{fig:1_our_method_1}
    \end{subfigure}
    \begin{subfigure}[b]{0.22\textwidth}
        \includegraphics[width=\textwidth]{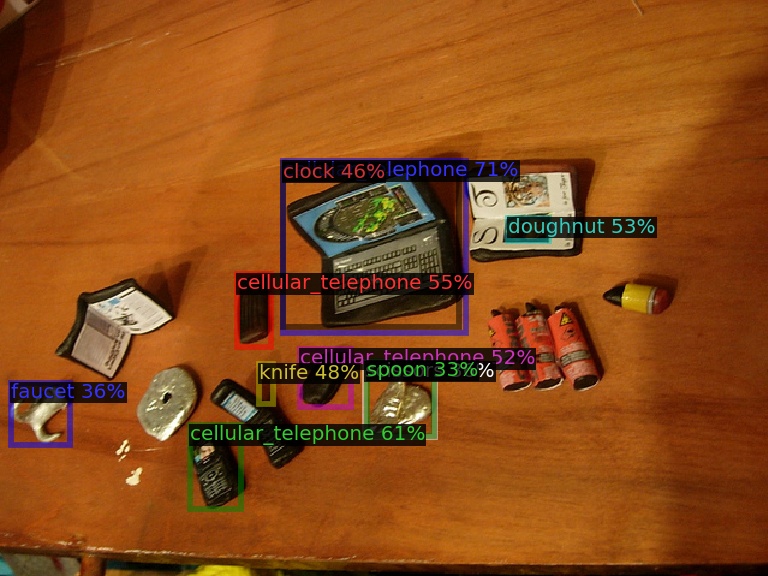}
        \label{fig:1_rncdl_1}
    \end{subfigure}
    \begin{subfigure}[b]{0.22\textwidth}
        \includegraphics[width=\textwidth]{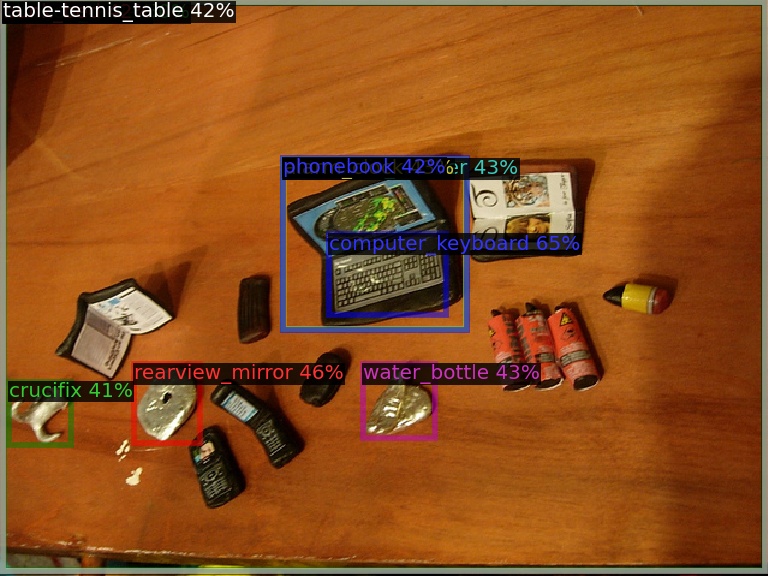}
        \label{fig:1_gdino_1}
    \end{subfigure}
    \begin{subfigure}[b]{0.22\textwidth}
        \includegraphics[width=\textwidth]{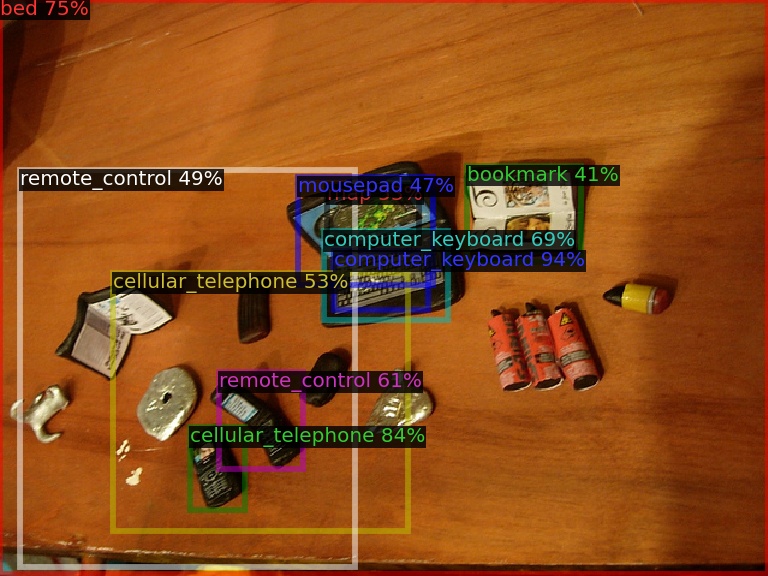}
        \label{fig:1_our_method_1}
    \end{subfigure}
    \begin{subfigure}[b]{0.22\textwidth}
        \includegraphics[width=\textwidth]{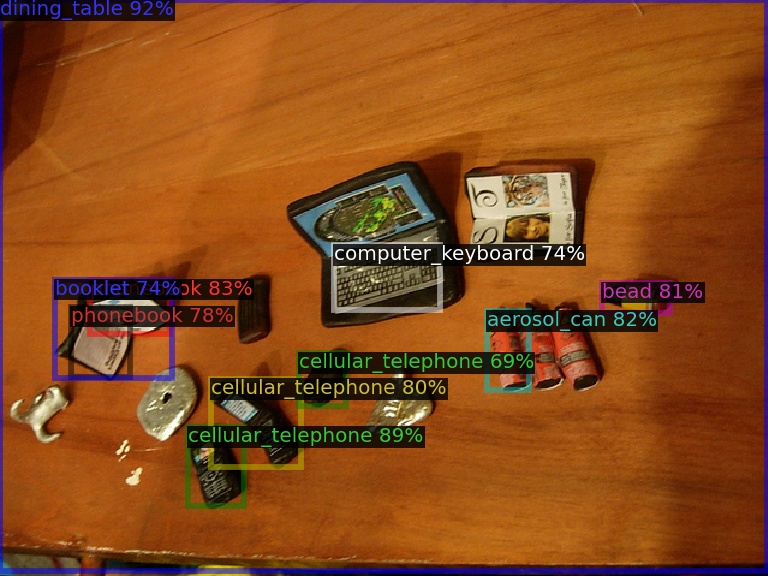}
        \label{fig:1_our_method_1}
    \end{subfigure}
    \begin{subfigure}[b]{0.22\textwidth}
        \includegraphics[width=\textwidth]{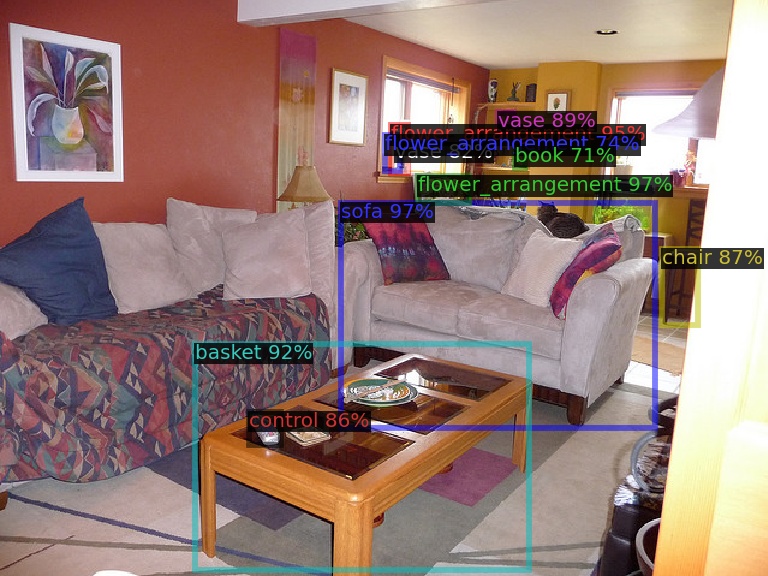}
        \caption*{(a)}
        \label{fig:2_rncdl}
    \end{subfigure}
    \begin{subfigure}[b]{0.22\textwidth}
        \includegraphics[width=\textwidth]{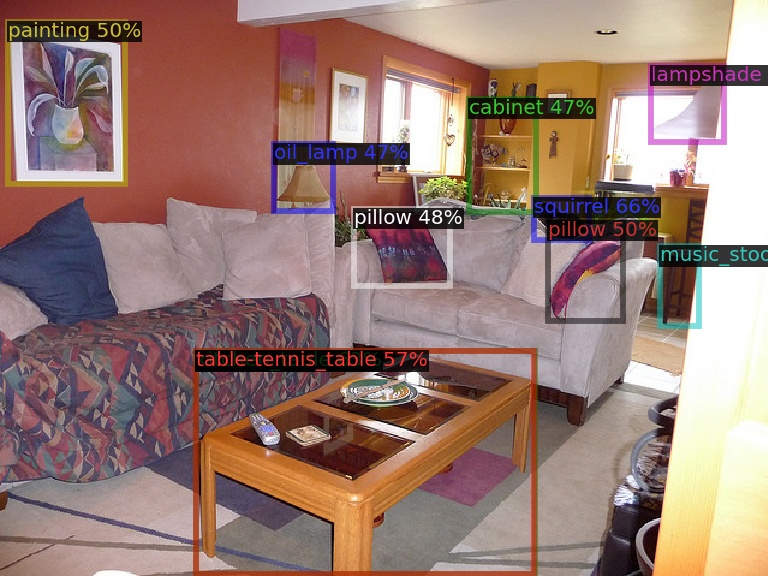}
        \caption*{(b)}
        \label{fig:2_gdino}
    \end{subfigure}
    \begin{subfigure}[b]{0.22\textwidth}
        \includegraphics[width=\textwidth]{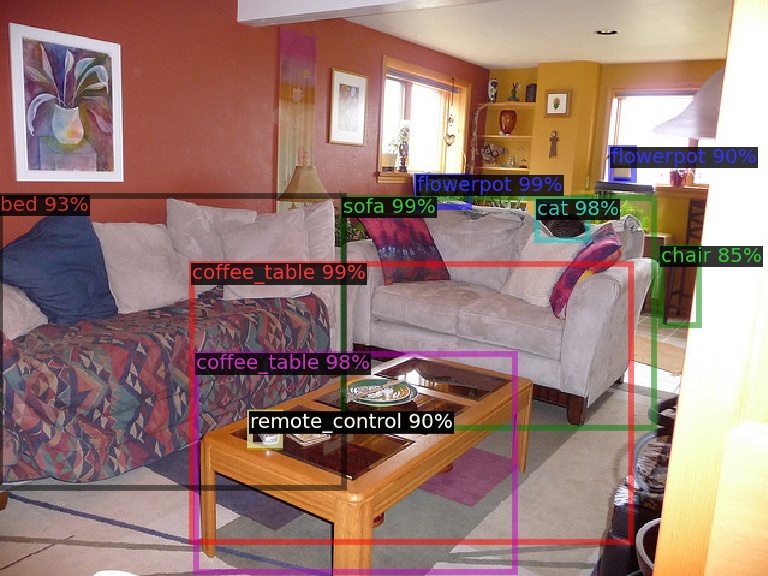}
        \caption*{(c)}
        \label{fig:2_our_method}
    \end{subfigure}
    \begin{subfigure}[b]{0.22\textwidth}
        \includegraphics[width=\textwidth]{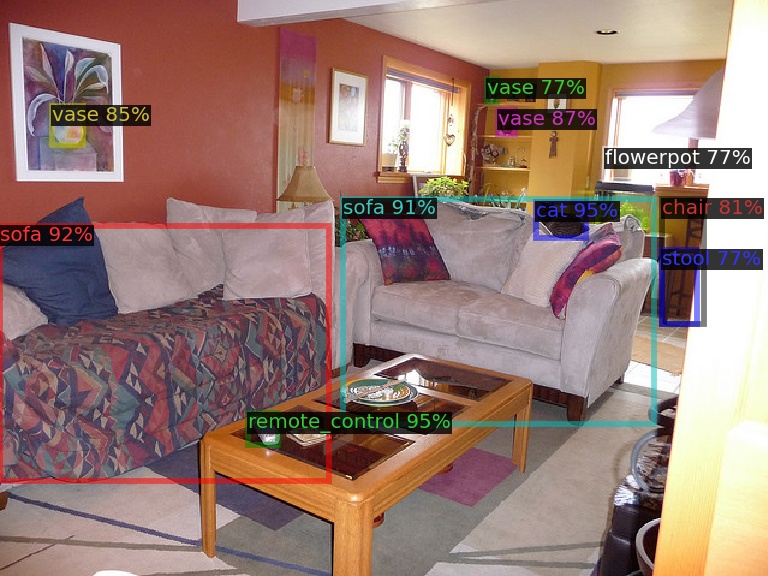}
        \caption*{(d)}
        \label{fig:2_our_method}
    \end{subfigure}
    \caption{Comparison of top-10 predictions by different NOD methods. RNCDL (column a) shows mislabelings, such as `\texttt{aerosol\_can}', `\texttt{doughnut}', `\texttt{basket}' and `\texttt{spoon}' due to post-discovery class assignment issues. GDINO (column b) has several inaccuracies, like `\texttt{grizzly}', `\texttt{wet\_suit}', `\texttt{rearview\_mirror}', and `\texttt{squirrel}'. It also produces a lot of uncertain and low confidence predictions (i.e. score $< 50\%$). Mask-RCNN with CLIP baseline (column c, without our refinement) misses out on objects like `\texttt{frying\_pan}', and outputs incorrect detections like `\texttt{bed}'. Our method (column d, GDINO+Mask-RCNN) accurately detects objects with high confidence and outperforms prior methods in precise localization and recognition of unique objects. Figure best viewed with zoom. }
\label{fig:qualitative_comparisons}
\end{figure*}

In \cref{sec:imple-setting}, we begin by outlining the implementation settings for our study.
Following this, in \cref{sec:main-results}, we present our main findings. We compare our work with the current leading models, GDINO~\cite{groundingdino}, and the RNCDL~\cite{fomenko2022learning}. 
Since our known and novel class splits on LVIS (80 known; 1123 novel) is more challenging than the conventional LVIS-OVD splits (866 known; 337 novel), we adapt our approach on the COCO OVD split for direct comparison against existing OVD methods in \cref{sec:open-vocab-results}. We also report comparisons on the localization capabilities of our method against the latest in \textit{unknown object detection}~\cite{liang2023unknown} and \textit{open-set object detection}~\cite{groundingdino} methods in \cref{sec:localization-results}. Our method shows a significant improvement over previous methods. In \cref{sec:ablation}, we report extensive ablation studies. The ablations demonstrate the valuable contributions of each component within our overall proposed approach. We conclude by showing the qualitative visualization of our method in \cref{fig:qualitative_comparisons}, showing clear improvements with our proposed cooperative mechanism.

\subsection{Implementation Settings}
\label{sec:imple-setting}
\textbf{Datasets:} In our experiments, we primarily utilize two datasets: COCO 2017~\cite{coco}, and LVIS v1~\cite{gupta2019lvis}. 
The LVIS dataset, while based on the same images from COCO, offers more detailed annotations. It features bounding box and instance segmentation mask annotations for 1,203 classes, encompassing all classes from COCO. 
The LVIS dataset is divided into 100K training images (\texttt{lvis\_train}) and 20K validation images (\texttt{lvis\_val}). Our principal evaluation, as detailed in \cref{sec:main-results}, focuses on the LVIS validation split, consistent with prior work~\cite{fomenko2022learning}. Additionally, we organize the LVIS validation images in descending order by the count of ground-truth box annotations. From this, we select a subset of 745 images (\texttt{lvis\_val\_subset}) approximately covering all LVIS classes. This subset is used in our ablation studies (\cref{sec:ablation}) and localization-only experiments (\cref{sec:localization-results}). In general, our investigations indicate that results on this subset are representative of those on the full validation set. \textbf{Known/Novel Split:} Consistent with RNCDL~\cite{fomenko2022learning}, we classify the 80 COCO classes as `known' and the remaining 1,123 LVIS classes as `novel'. To align with the COCO OVD split, we categorize 48 COCO classes as known and the rest of the 17 as novel. 

\begin{table}[!t]
    \centering
    \resizebox{\columnwidth}{!}{%
    \begin{tabular}{lcccccc}
        \toprule
        Method & Mask-RCNN & GDINO & VLM & Novel AP & Known AP & All AP \\
        \midrule
        K-Means~\cite{kmeans} & - & - & - & 0.20 & 17.77 & 1.55 \\
        Weng et al.~\cite{weng2021unsupervised} & - & - & - & 0.27 & 17.85 & 1.62 \\
        ORCA~\cite{orca} & - & - & - & 0.49 & 20.57 & 2.03 \\
        UNO~\cite{UNO} & - & - & - & 0.61 & 21.09 & 2.18 \\
        RNCDL~\cite{fomenko2022learning} & V1 & - & - & 5.42 & 25.00 & 6.92 \\
        GDINO~\cite{groundingdino} & - & \cmark & - & 13.47 & 37.13 & 15.30 \\
        \rowcolor{gray!30}
        \textbf{Ours} & V2 & \xmark & CLIP & 13.24 & \textbf{42.61} & 15.52 \\
        \rowcolor{gray!30}
        \textbf{Ours} & V1 & \cmark & CLIP & 15.37 & 36.15 & 16.98 \\
        \rowcolor{gray!30}
        \textbf{Ours} & V1 & \cmark & SigLIP & 16.12 & 37.09 & 17.74 \\
        \rowcolor{gray!30}
        \textbf{Ours} & V2 & \cmark & SigLIP & \textbf{17.42} & 42.08 & \textbf{19.33} \\
        \bottomrule
    \end{tabular}%
    }
    \caption{Comparison of object detection performance using mAP on the \texttt{lvis\_val} dataset. The VLM column specifies the Vision-Language Model (VLM) utilized, indicating whether it is CLIP or SigLIP. Overall, the SigLIP version with GDINO and Mask-RCNN-V2 provides the best novel vs. known AP tradeoff.}
    \label{tab:performance-comparison}
\end{table}

\noindent \textbf{Mask-RCNN:} We explore two versions of the pre-trained Mask-RCNN model. The first, referred to as ``\textit{Mask-RCNN-V1}'', adheres to the fully-supervised training methodology of RNCDL~\cite{fomenko2022learning}, trained on the \( \text{COCO}_{\text{half}} \) dataset to ensure fair comparison. The second version, ``\textit{Mask-RCNN-V2}'', is trained on the complete COCO dataset, utilizing a ResNet101~\cite{resnet} based FPN~\cite{fpn} backbone with large scale jittering (LSJ)~\cite{lsj} augmentation. We consider top 300 predictions from Mask-RCNN (i.e. \( N_{\text{KN}} + N_{\text{BG}} = 300\)).
\textbf{GDINO:} For our experiments, the Swin-T variant of GDINO is employed, being the only open-sourced variant compatible with our open-vocabulary novel-class set constraint. The \texttt{num\_query} parameter in GDINO is set to 900, but only the top-300 high-scoring predictions (\( N_{\text{GD}} = 300 \)) are utilized (\cref{fig:box-ablation}). 
\textbf{CLIP:} Two variants of CLIP-based models are utilized in our approach. The first is the standard CLIP model with a ``\texttt{ViT-L/14}'' backbone~\cite{clip}, and the second is the recently released SigLIP model~\cite{SigLIP, rw2019timm, openclip}. We use the ``\texttt{ViT-SO400M-14-SigLIP}'' backbone from SigLIP, alongside 64 prompt templates from ViLD~\cite{gu2022openvocabulary}, and synonyms from the LVIS dataset~\cite{gupta2019lvis} for our methodology. 

\noindent \textbf{Evaluation Metrics:} For the the principal results in \cref{sec:main-results}, and the ablation studies in \cref{sec:ablation}, we align with previous studies and report mean average precision (mAP@[0.5:0.95])~\cite{coco} across ``known'', ``novel'', and ``all'' classes. Further, following the LVIS protocol~\cite{gupta2019lvis}, we evaluate our method by finally taking top 300 high scoring predictions. For COCO OVD results in \cref{sec:open-vocab-results}, we evaluate by taking top 100 high scoring predictions. In \cref{sec:localization-results}, recall values at an IOU threshold of 0.5 are evaluated for the localization experiments. For the open-vocabulary comparison in \cref{sec:open-vocab-results}, box \( \text{AP}_{50} \) metrics are reported, following the precedent set by earlier works.

\subsection{Comparison with NOD Techniques}
\label{sec:main-results}

\begin{table}[htbp]
\centering
\resizebox{\columnwidth}{!}{%
\begin{tabular}{p{2cm} p{3.5cm} p{3.5cm} p{2cm} ccc}
\toprule
Method & Pre-train & Training & Backbone & \textbf{Novel} & \textcolor{gray}{\textbf{Base}} & \textcolor{gray}{\textbf{All}} \\
\midrule
OVR-CNN~\cite{ovr-cnn} & COCO Captions, BERT (BooksCorpus, English Wikipedia) & COCO & ResNet50 & 22.8 & \textcolor{gray}{46.0} & \textcolor{gray}{39.9} \\
ViLD~\cite{gu2022openvocabulary} & - & COCO, CLIP & ResNet50 & 27.6 & \textcolor{gray}{59.5} & \textcolor{gray}{51.3} \\
Detic~\cite{detic} & ImageNet-21K & COCO, ImageNet-21K, Conceptual Captions, CLIP & ResNet50 & 27.8 & \textcolor{gray}{47.1} & \textcolor{gray}{45.0} \\
OV-DETR~\cite{ov-detr} & - & COCO, CLIP & ResNet50-C4 & 29.4 & \textcolor{gray}{61.0} & \textcolor{gray}{52.7} \\
BARON~\cite{wu2023aligning} & SOCO dataset & COCO, CLIP & ResNet50 & 34 & \textcolor{gray}{60.4} & \textcolor{gray}{53.5} \\
CORA~\cite{wu2023cora} & - & COCO, CLIP & ResNet50 & 35.1 & \textcolor{gray}{35.5} & \textcolor{gray}{35.4} \\
Rasheed et al.~\cite{rasheed2022bridging} & MAVL (Flickr30k, COCO, Visual Genome) & COCO, COCO Captions, CLIP & ResNet50 & 36.6 & \textcolor{gray}{54.0} & \textcolor{gray}{49.4} \\
BARON~\cite{wu2023aligning} & SOCO dataset, MAVL & COCO, COCO Captions, CLIP & ResNet50-C4 & 42.7 & \textcolor{gray}{54.9} & \textcolor{gray}{51.7} \\
CORA+~\cite{wu2023cora} & - & COCO, COCO Captions, CLIP & ResNet50x4 & 43.1 & \textcolor{gray}{60.9} & \textcolor{gray}{56.2} \\
DetCLIPv3~\cite{yao2024detclipv3} & FILIP, Qformer, BERT, CLIP & O365, GoldG, V3Det, GranuCap50M, GranuCap600K, CLIP, InstructBLIP, GPT-4, LLaVA & Swin-T & 54.7 & \textcolor{gray}{42.8} & \textcolor{gray}{46.9} \\
\addlinespace
\rowcolor{gray!30}
\textbf{Ours*} & \textbf{GDINO (O365,GoldG,Cap4M), SAM (SA-1B), CLIP} & COCO & \textbf{ResNet50} & \textbf{49.6} & \textcolor{gray}{42.4} & \textcolor{gray}{44.3} \\
\addlinespace
\rowcolor{gray!30}
\textbf{Ours*} & \textbf{GDINO (O365,GoldG,Cap4M), SAM (SA-1B), CLIP} & COCO & \textbf{ResNet101} & \textbf{50.3} & \textcolor{gray}{49.8} & \textcolor{gray}{49.9} \\
\bottomrule
\end{tabular}%
}
\caption{Comparison with OVD methods on COCO-OVD data split. ``COCO'' in the training column refers to the COCO dataset with 48 base class annotations. Our method requires only the closed-set detector to be trained or can be training-free with pre-trained detectors, unlike most existing OVD methods that require explicit training with CLIP or other VLMs. Our method is thus zero-shot, while OVD methods are not (evident from CLIP being present in training column of most methods).}
\label{tab:method_comparison}
\end{table}

The results in \cref{tab:performance-comparison} reveal that baseline methods like K-Means and other GNCD approaches demonstrate limited effectiveness in NOD (Novel Object Detection), with Novel AP below 1 mAP. Even RNCDL~\cite{fomenko2022learning}, which involves a sophisticated multi-stage training process and requires extensive hyperparameter tuning, falls short in terms of both Known and Novel AP compared to the GDINO benchmark.
In stark contrast, our method, which does not necessitate any training and is conceptually straightforward, significantly outperforms RNCDL. Specifically, with our method using only CLIP, we achieve an improvement of 7.82 mAP in Novel classes and an impressive 17.61 mAP in Known classes over RNCDL. Moreover, by integrating our cooperative mechanism with the GDINO and SigLIP model, we observe additional gains of 3.95 mAP in Novel classes and 4.95 mAP in Known classes compared to the baseline GDINO. The use of advanced VLMs like SigLIP and enhanced pre-trained weights of Mask-RCNN further bolsters the overall object detection performance. This showcases the scalability and modular nature of our approach, reinforcing its potential for diverse NOD applications.

\subsection{Comparisons with OVD Methods}
\label{sec:open-vocab-results}
\cref{tab:method_comparison} compares our proposed zero-shot approach against various SOTA OVD methods on the COCO-OVD dataset split. Unlike most OVD approaches that require extensive training with VLMs like CLIP, our method leverages only a pre-trained closed-set detector and uses VLMs like CLIP in zero-shot way. Specifically, our method achieves a notable 49.6\% AP50 on novel categories, surpassing the majority of the existing methods, including those that rely heavily on VLM training such as CORA+ (43.1\%) and BARON MAVL (42.7\%). While DetCLIPv3 demonstrates a higher Novel AP50 of 54.7\%, this comparison is not entirely fair due to the substantial pre-training and training resources it employs (e.g., FILIP, Qformer, GPT-4, and others). While our method uses multiple models during pre-training, it does not use any such model in training (except detector). DetCLIPv3 uses models in both pre-training and training. Due to the zero-shot nature of our method, NOD benchmark in \cref{tab:performance-comparison} becomes the main result.

\begin{figure}[!t]
    \centering
    \includegraphics[width=0.9\columnwidth]{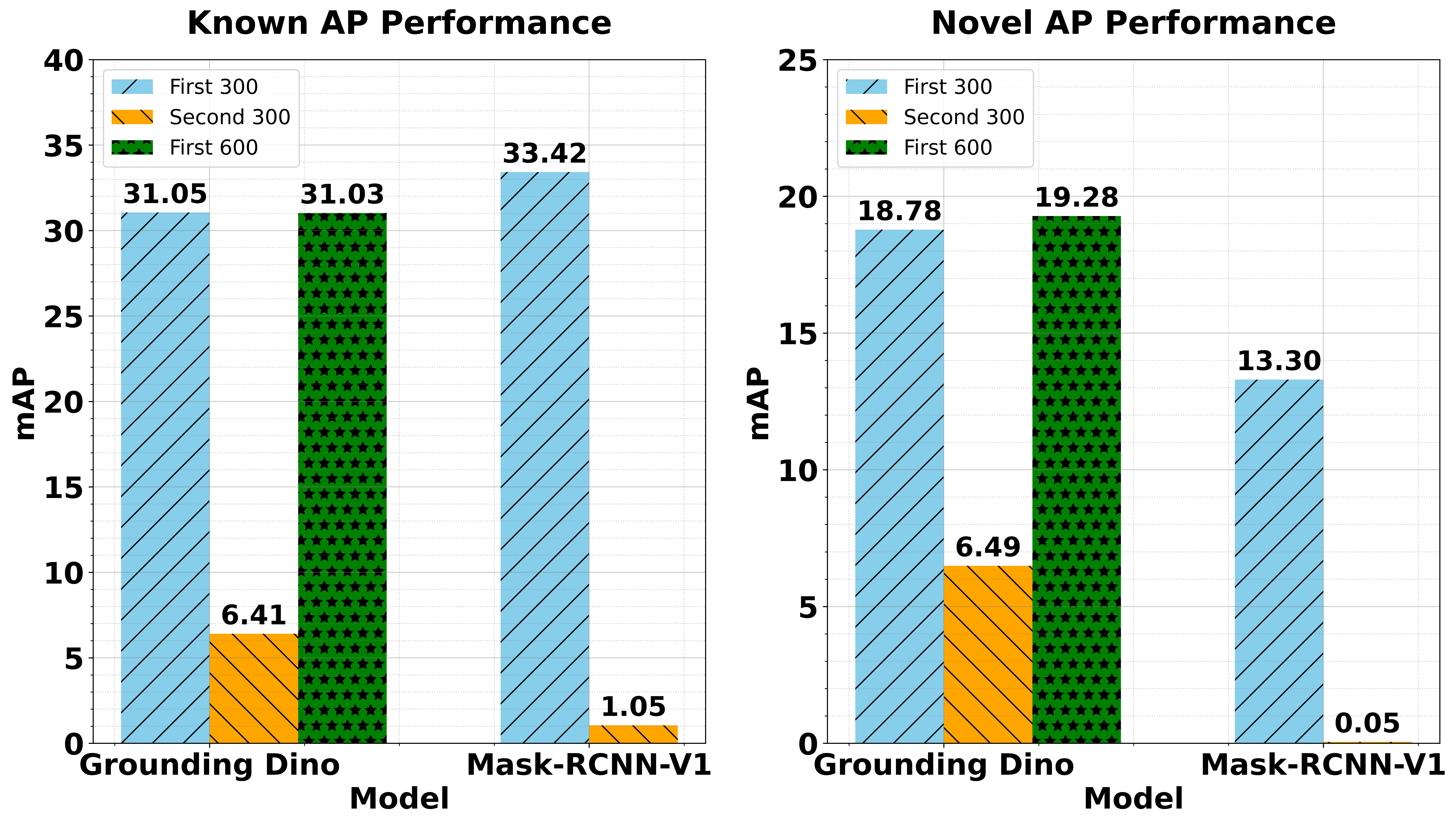}
    \caption{Object detection ``mAP'' performance on the \texttt{lvis\_val\_subset} dataset. ``Grounding Dino'' and ``Mask-RCNN-V1'' models were evaluated after considering the first 300, second 300, and the first 600 boxes.}
    \label{fig:box-ablation}
\end{figure}

\begin{figure}[t]
    \centering
    \includegraphics[width=0.85\columnwidth]{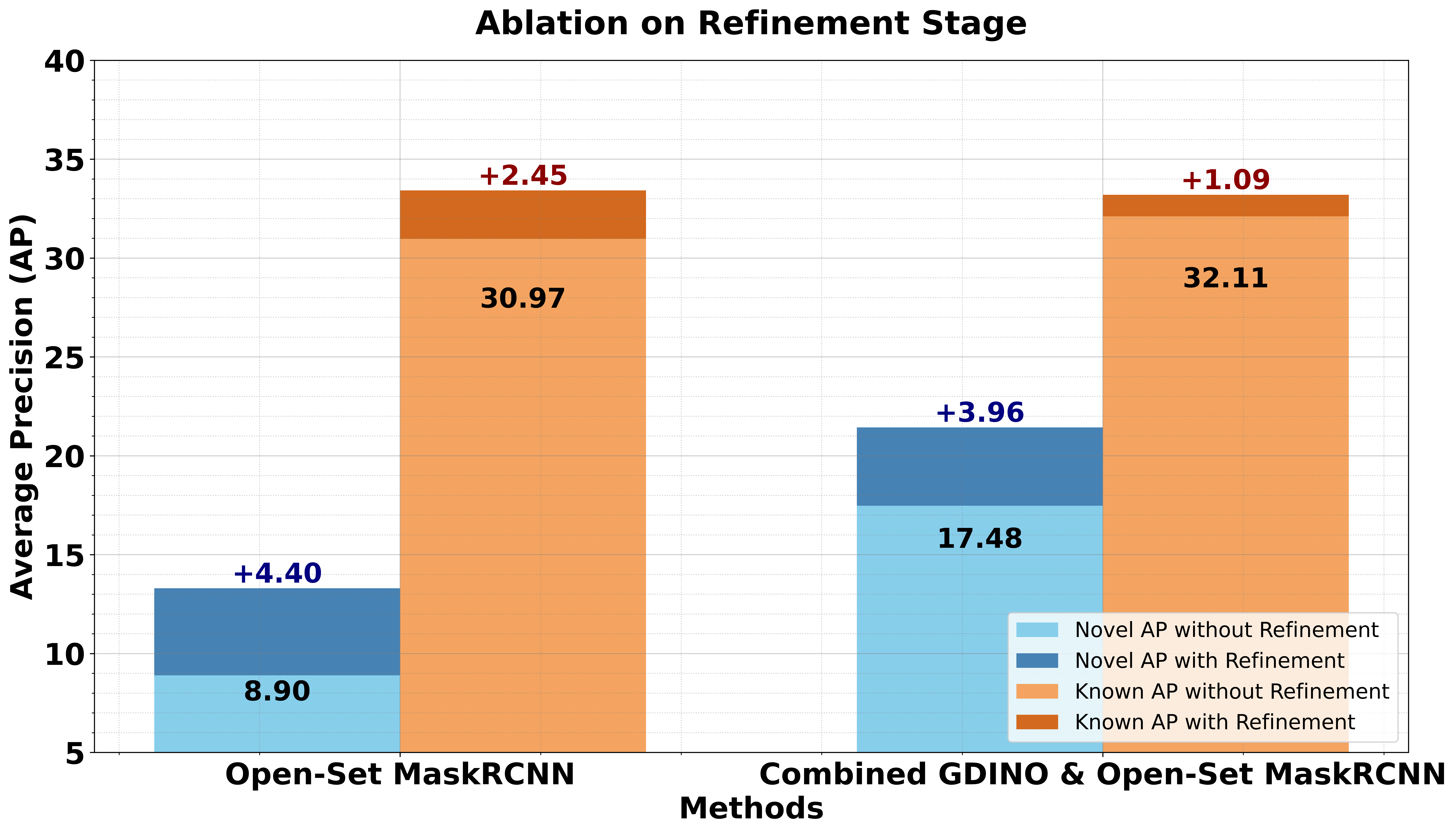}
    \caption{Ablation of our refinement component on the \texttt{lvis\_val\_subset} dataset. The figure reports the Average Precision (AP) for both Novel and Known objects.}
    \label{fig:ablation-refinement}
\end{figure}

\subsection{Comparisons with SOTA Unknown Object  Detection Methods}
\label{sec:localization-results}
\cref{tab:localization} presents a comparison between our approach and the SOTA in unknown object detection, specifically the UnSniffer model \cite{liang2023unknown}. To align with our evaluation criteria, we retrained UnSniffer on the entire set of COCO classes, excluding its NCut filtering feature for fair comparison. Our results demonstrate a notable improvement over existing methods, such as RNCDL, GDINO, and UnSniffer, in accurately localizing novel and known objects. Our method achieves a leading recall rate of 37.36\%, showing an absolute 3.5\% gain over the second best method, RNCDL.

\begin{table}[t]
  \centering
  \resizebox{0.75\columnwidth}{!}{%
  \begin{tabular}{lcccc}
    \toprule
    Method          & Recall (\%) & Num TP & Num GT & Tot Pred \\
    \midrule
    GDINO~\cite{groundingdino}           & 29.96       & 13653  & 45570  & 223500   \\
    UnSniffer~\cite{liang2023unknown}       & 30.83       & 14048  & 45570  & 216424   \\
    RNCDL~\cite{fomenko2022learning}   & 33.82       & 15413  & 45570  & 216553   \\
    \rowcolor{gray!30}
    \textbf{Ours*} & \textbf{37.36}       & \textbf{17026}  & 45570  & 223500   \\
    \bottomrule
  \end{tabular}%
  }
  \caption{Comparison of localization performance of different methods on the \texttt{lvis\_val\_subset} dataset. Recall is measured with an IOU threshold of 0.5. True positives (TP) are those predicted boxes with IOU $>$ 0.5 with any ground truth (GT) box. *Our method with GDINO, CLIP, and Mask-RCNN-V1.}
  \label{tab:localization}
\end{table}

\begin{table}[t]
    \centering\setlength{\tabcolsep}{2pt}
    \resizebox{\columnwidth}{!}{%
    \begin{tabular}{lcccccccc}
        \toprule
        Method & SigLIP & SAM & GDINO & SRM & SAEG & AP (Novel) & AP (Known) & AP (All) \\
        \midrule
        Ours & \xmark & \cmark & \cmark & \cmark & \cmark & 7.46 & \textbf{39.44} & 9.93 \\
        Ours & \cmark & \xmark & \cmark & \cmark & \cmark & 11.54 & 34.62 & 13.32 \\
        Ours & \cmark & \cmark & \xmark & \cmark & \cmark & 11.39 & 36.81 & 13.35 \\
        Ours & \cmark & \cmark & \cmark & \xmark & \cmark & 14.32 & 36.66 & 16.05 \\
        Ours & \cmark & \cmark & \cmark & \cmark & \xmark & 15.84 & 37.07 & 17.48 \\
        \rowcolor{gray!30}
        \textbf{Ours} & \cmark & \cmark & \cmark & \cmark & \cmark & \textbf{16.12} & 37.09 & \textbf{17.74} \\
        \bottomrule
    \end{tabular}%
    }
    \caption{Ablation experiments of our method consisting of {SigLIP + GDINO + Mask-RCNN-V1} and our proposed components on the \texttt{lvis\_val} set. Each row represents the performance of the final method minus a particular component indicated by the \xmark.}
    \label{tab:ablation-experiments-siglip}
\end{table}

\subsection{Ablations}
\label{sec:ablation}

\textbf{Ablation on Top-N Box Predictions - Grounding DINO vs. Mask-RCNN:} We examine the mAP performance of GDINO and Mask-RCNN for both known and novel classes with different numbers of box predictions (\cref{fig:box-ablation}). For GDINO, increasing \( N_{\text{GD}} \) beyond 300 leads to only marginal performance improvements, whereas considering the first 600 boxes leads to only 0.5 mAP increase on novel classes. Considering this alongside GPU memory constraints, we decide to set \( N_{\text{GD}} \) at 300 for optimal efficiency. In the case of Mask-RCNN, as we can see from \cref{fig:box-ablation}, the known AP considering the next 300 boxes is only 1.05 mAP, while the novel AP is mere 0.05 mAP, therefore the top-300 predictions (\( N_{\text{KN}} + N_{\text{BG}} = 300 \)) is optimal, and further predictions introduce more noise and do not significantly capture ground-truth generic objects.

\textbf{Ablation on Refinement Stage:} Our ablation study on the refinement stage (\cref{fig:ablation-refinement}) shows significant improvements due to proposed refinement process. The refinement stage added to open-set Mask RCNN results in a notable 4.4 mAP gain on novel AP. Further, combining GDINO and open-set Mask-RCNN predictions with our refinement stage yields superior performance, resulting in gains of 3.96 mAP in novel, and 1.09 mAP in known classes.

\textbf{Ablation of Components in our Cooperative Mechanism:}
Our experiments assess the impact of each component in our method by observing changes in mAP when a component is removed. As indicated in \cref{tab:ablation-experiments-siglip}, excluding SigLIP causes the most significant drop in Novel AP (8.66 mAP), suggesting its crucial role in detecting novel classes. Conversely, this removal slightly improves known AP by 2.35 mAP, highlighting a trade-off between novel and known AP detection. The absence of GDINO reduces novel AP by 4.73 mAP, confirming its importance in identifying novel objects. Similarly, without SAM, there is a 4.58 mAP decrease in novel AP. This implies that even with GDINO and SigLIP, optimal performance on novel classes requires the refinement offered by SAM (\cref{subsec:Refinement}). Additionally, the performance across both known and novel classes drops significantly without our novel components (SAEG and SRM), demonstrating their effectiveness.

\section{Conclusion}
In this research, we introduce a novel cooperative mechanism that leverages the complementary strengths of pre-trained foundational models. 
This mechanism effectively transforms any existing closed-set detector into an open-set detector, addressing the challenges of novel object detection (NOD). The modular design of our approach allows for seamless integration with current open-set detectors, enhancing their performance in detecting both known and novel classes.
Our method has consistently outperforms state-of-the-art techniques in various settings, including NOD, unknown object localization, and open-vocabulary detection.
While our study primarily focused on bounding box outputs, the underlying components, namely Mask-RCNN and SAM, are inherently capable of generating instance segmentation masks. This aspect suggests that our methodology can be readily extended to instance segmentation tasks, indicating a broader scope of application.
Overall, the cooperative mechanism we propose not only addresses current limitations in object detection for NOD but also provides a versatile and effective framework that can be deployed in other applications.

\noindent \textbf{Limitations:} Despite the superior performance of our proposed method in NOD, when compared to existing approaches, a primary limitation is the inference speed and potential data leakage. Addressing this limitation (by training efficient custom models from scratch) without compromising its detection capabilities is a key area for future research.

{\small
\bibliographystyle{ieee_fullname}
\bibliography{main}

\begin{thebibliography}{10}\itemsep=-1pt

\bibitem{orca}
Kaidi Cao, Maria Brbic, and Jure Leskovec.
\newblock Open-world semi-supervised learning.
\newblock {\em arXiv preprint arXiv:2102.03526}, 2021.

\bibitem{detr}
Nicolas Carion, Francisco Massa, Gabriel Synnaeve, Nicolas Usunier, Alexander Kirillov, and Sergey Zagoruyko.
\newblock End-to-end object detection with transformers, 2020.

\bibitem{driving-autono}
Chenyi Chen, Ari Seff, Alain Kornhauser, and Jianxiong Xiao.
\newblock Deepdriving: Learning affordance for direct perception in autonomous driving.
\newblock In {\em 2015 IEEE International Conference on Computer Vision (ICCV)}, pages 2722--2730, 2015.

\bibitem{bert}
Jacob Devlin, Ming-Wei Chang, Kenton Lee, and Kristina Toutanova.
\newblock Bert: Pre-training of deep bidirectional transformers for language understanding, 2019.

\bibitem{du2022vos}
Xuefeng Du, Zhaoning Wang, Mu Cai, and Yixuan Li.
\newblock Vos: Learning what you don't know by virtual outlier synthesis, 2022.

\bibitem{UNO}
Enrico Fini, Enver Sangineto, Stéphane Lathuilière, Zhun Zhong, Moin Nabi, and Elisa Ricci.
\newblock A unified objective for novel class discovery, 2021.

\bibitem{fomenko2022learning}
Vladimir Fomenko, Ismail Elezi, Deva Ramanan, Laura Leal-Taix{'e}, and Aljo\v{s}a O\v{s}ep.
\newblock Learning to discover and detect objects.
\newblock In {\em Advances in Neural Information Processing Systems}, 2022.

\bibitem{lsj}
Golnaz Ghiasi, Yin Cui, Aravind Srinivas, Rui Qian, Tsung-Yi Lin, Ekin~D. Cubuk, Quoc~V. Le, and Barret Zoph.
\newblock Simple copy-paste is a strong data augmentation method for instance segmentation, 2021.

\bibitem{gu2022openvocabulary}
Xiuye Gu, Tsung-Yi Lin, Weicheng Kuo, and Yin Cui.
\newblock Open-vocabulary object detection via vision and language knowledge distillation, 2022.

\bibitem{gupta2019lvis}
Agrim Gupta, Piotr Dollár, and Ross Girshick.
\newblock Lvis: A dataset for large vocabulary instance segmentation, 2019.

\bibitem{maskrcnn}
Kaiming He, Georgia Gkioxari, Piotr Dollár, and Ross Girshick.
\newblock Mask r-cnn.
\newblock In {\em 2017 IEEE International Conference on Computer Vision (ICCV)}, pages 2980--2988, 2017.

\bibitem{resnet}
Kaiming He, Xiangyu Zhang, Shaoqing Ren, and Jian Sun.
\newblock Deep residual learning for image recognition.
\newblock In {\em 2016 IEEE Conference on Computer Vision and Pattern Recognition (CVPR)}, pages 770--778, 2016.

\bibitem{openclip}
Gabriel Ilharco, Mitchell Wortsman, Ross Wightman, Cade Gordon, Nicholas Carlini, Rohan Taori, Achal Dave, Vaishaal Shankar, Hongseok Namkoong, John Miller, Hannaneh Hajishirzi, Ali Farhadi, and Ludwig Schmidt.
\newblock Openclip, July 2021.
\newblock If you use this software, please cite it as below.

\bibitem{sam}
Alexander Kirillov, Eric Mintun, Nikhila Ravi, Hanzi Mao, Chloe Rolland, Laura Gustafson, Tete Xiao, Spencer Whitehead, Alexander~C. Berg, Wan-Yen Lo, Piotr Dollár, and Ross Girshick.
\newblock Segment anything, 2023.

\bibitem{li2021grounded}
Liunian~Harold Li*, Pengchuan Zhang*, Haotian Zhang*, Jianwei Yang, Chunyuan Li, Yiwu Zhong, Lijuan Wang, Lu Yuan, Lei Zhang, Jenq-Neng Hwang, Kai-Wei Chang, and Jianfeng Gao.
\newblock Grounded language-image pre-training.
\newblock In {\em CVPR}, 2022.

\bibitem{liang2023unknown}
Wenteng Liang, Feng Xue, Yihao Liu, Guofeng Zhong, and Anlong Ming.
\newblock Unknown sniffer for object detection: Don't turn a blind eye to unknown objects.
\newblock In {\em Proceedings of the IEEE/CVF Conference on Computer Vision and Pattern Recognition}, pages 3230--3239, 2023.

\bibitem{fpn}
Tsung-Yi Lin, Piotr Dollár, Ross Girshick, Kaiming He, Bharath Hariharan, and Serge Belongie.
\newblock Feature pyramid networks for object detection.
\newblock In {\em 2017 IEEE Conference on Computer Vision and Pattern Recognition (CVPR)}, pages 936--944, 2017.

\bibitem{coco}
Tsung-Yi Lin, Michael Maire, Serge Belongie, Lubomir Bourdev, Ross Girshick, James Hays, Pietro Perona, Deva Ramanan, C.~Lawrence Zitnick, and Piotr Dollár.
\newblock Microsoft coco: Common objects in context, 2015.

\bibitem{groundingdino}
Shilong Liu, Zhaoyang Zeng, Tianhe Ren, Feng Li, Hao Zhang, Jie Yang, Chunyuan Li, Jianwei Yang, Hang Su, Jun Zhu, et~al.
\newblock Grounding dino: Marrying dino with grounded pre-training for open-set object detection.
\newblock {\em arXiv preprint arXiv:2303.05499}, 2023.

\bibitem{swintrans}
Ze Liu, Yutong Lin, Yue Cao, Han Hu, Yixuan Wei, Zheng Zhang, Stephen Lin, and Baining Guo.
\newblock Swin transformer: Hierarchical vision transformer using shifted windows, 2021.

\bibitem{kmeans}
James MacQueen et~al.
\newblock Some methods for classification and analysis of multivariate observations.
\newblock In {\em Proceedings of the fifth Berkeley symposium on mathematical statistics and probability}, volume~1, pages 281--297. Oakland, CA, USA, 1967.

\bibitem{patterson2021carbon}
David Patterson, Joseph Gonzalez, Quoc Le, Chen Liang, Lluis-Miquel Munguia, Daniel Rothchild, David So, Maud Texier, and Jeff Dean.
\newblock Carbon emissions and large neural network training, 2021.

\bibitem{scikit-learn}
F. Pedregosa, G. Varoquaux, A. Gramfort, V. Michel, B. Thirion, O. Grisel, M. Blondel, P. Prettenhofer, R. Weiss, V. Dubourg, J. Vanderplas, A. Passos, D. Cournapeau, M. Brucher, M. Perrot, and E. Duchesnay.
\newblock Scikit-learn: Machine learning in {P}ython.
\newblock {\em Journal of Machine Learning Research}, 12:2825--2830, 2011.

\bibitem{clip}
Alec Radford, Jong~Wook Kim, Chris Hallacy, Aditya Ramesh, Gabriel Goh, Sandhini Agarwal, Girish Sastry, Amanda Askell, Pamela Mishkin, Jack Clark, Gretchen Krueger, and Ilya Sutskever.
\newblock Learning transferable visual models from natural language supervision, 2021.

\bibitem{rasheed2022bridging}
Hanoona Rasheed, Muhammad Maaz, Muhammad~Uzair Khattak, Salman Khan, and Fahad~Shahbaz Khan.
\newblock Bridging the gap between object and image-level representations for open-vocabulary detection, 2022.

\bibitem{rizve2022openldn}
Mamshad~Nayeem Rizve, Navid Kardan, Salman Khan, Fahad Shahbaz~Khan, and Mubarak Shah.
\newblock Openldn: Learning to discover novel classes for open-world semi-supervised learning.
\newblock In {\em European Conference on Computer Vision}, pages 382--401. Springer, 2022.

\bibitem{strubell2020energy}
Emma Strubell, Ananya Ganesh, and Andrew McCallum.
\newblock Energy and policy considerations for modern deep learning research.
\newblock In {\em Proceedings of the AAAI conference on artificial intelligence}, volume~34, pages 13693--13696, 2020.

\bibitem{vaze2022generalized}
Sagar Vaze, Kai Han, Andrea Vedaldi, and Andrew Zisserman.
\newblock Generalized category discovery.
\newblock In {\em Proceedings of the IEEE/CVF Conference on Computer Vision and Pattern Recognition}, pages 7492--7501, 2022.

\bibitem{wang2023multi}
Yingjie Wang, Qiuyu Mao, Hanqi Zhu, Jiajun Deng, Yu Zhang, Jianmin Ji, Houqiang Li, and Yanyong Zhang.
\newblock Multi-modal 3d object detection in autonomous driving: a survey.
\newblock {\em International Journal of Computer Vision}, pages 1--31, 2023.

\bibitem{weng2021unsupervised}
Zhenzhen Weng, Mehmet~Giray Ogut, Shai Limonchik, and Serena Yeung.
\newblock Unsupervised discovery of the long-tail in instance segmentation using hierarchical self-supervision, 2021.

\bibitem{rw2019timm}
Ross Wightman.
\newblock Pytorch image models.
\newblock \url{https://github.com/rwightman/pytorch-image-models}, 2019.

\bibitem{wu2023aligning}
Size Wu, Wenwei Zhang, Sheng Jin, Wentao Liu, and Chen~Change Loy.
\newblock Aligning bag of regions for open-vocabulary object detection, 2023.

\bibitem{wu2023cora}
Xiaoshi Wu, Feng Zhu, Rui Zhao, and Hongsheng Li.
\newblock Cora: Adapting clip for open-vocabulary detection with region prompting and anchor pre-matching, 2023.

\bibitem{yao2024detclipv3}
Lewei Yao, Renjie Pi, Jianhua Han, Xiaodan Liang, Hang Xu, Wei Zhang, Zhenguo Li, and Dan Xu.
\newblock Detclipv3: Towards versatile generative open-vocabulary object detection.
\newblock In {\em Proceedings of the IEEE/CVF Conference on Computer Vision and Pattern Recognition}, pages 27391--27401, 2024.

\bibitem{ov-detr}
Yuhang Zang, Wei Li, Kaiyang Zhou, Chen Huang, and Chen~Change Loy.
\newblock Open-vocabulary detr with conditional matching.
\newblock In Shai Avidan, Gabriel Brostow, Moustapha Ciss{\'e}, Giovanni~Maria Farinella, and Tal Hassner, editors, {\em Computer Vision -- ECCV 2022}, pages 106--122, Cham, 2022. Springer Nature Switzerland.

\bibitem{ovr-cnn}
Alireza Zareian, Kevin~Dela Rosa, Derek~Hao Hu, and Shih-Fu Chang.
\newblock Open-vocabulary object detection using captions, 2021.

\bibitem{SigLIP}
Xiaohua Zhai, Basil Mustafa, Alexander Kolesnikov, and Lucas Beyer.
\newblock Sigmoid loss for language image pre-training, 2023.

\bibitem{dino}
Hao Zhang, Feng Li, Shilong Liu, Lei Zhang, Hang Su, Jun Zhu, Lionel~M. Ni, and Heung-Yeung Shum.
\newblock Dino: Detr with improved denoising anchor boxes for end-to-end object detection, 2022.

\bibitem{zhang2023promptcal}
Sheng Zhang, Salman Khan, Zhiqiang Shen, Muzammal Naseer, Guangyi Chen, and Fahad~Shahbaz Khan.
\newblock Promptcal: Contrastive affinity learning via auxiliary prompts for generalized novel category discovery.
\newblock In {\em Proceedings of the IEEE/CVF Conference on Computer Vision and Pattern Recognition}, pages 3479--3488, 2023.

\bibitem{detic}
Xingyi Zhou, Rohit Girdhar, Armand Joulin, Philipp Krähenbühl, and Ishan Misra.
\newblock Detecting twenty-thousand classes using image-level supervision, 2022.

\bibitem{od-survey}
Zhengxia Zou, Keyan Chen, Zhenwei Shi, Yuhong Guo, and Jieping Ye.
\newblock Object detection in 20 years: A survey.
\newblock {\em Proceedings of the IEEE}, 111(3):257--276, 2023.

\end{thebibliography}
}

\end{document}